\documentclass[conference]{IEEEtran}

\usepackage{graphicx}
\usepackage{array}
\usepackage{url}
\usepackage[numbers]{natbib}

\title{A Layered View of LLM Support in Underwater Robotics}

\author{%
\IEEEauthorblockN{Federico Celi, Leonardo Zacchini, and Gabriele Ferri}
\IEEEauthorblockA{NATO Centre for Maritime Research and Experimentation, La Spezia, Italy\\
\{name.lastname\}@cmre.nato.int}
}

\begin{document}

\maketitle

\begin{abstract}
This extended abstract discusses a specific and, in our experience, practically relevant role for large language models (LLMs) in underwater robotics. Our main premise is that, in underwater fielding missions, the immediate difficulty is often not the synthesis of new low-level control laws, but the operation of complex robotic systems under limited communications, partial observability, and expensive field time. In particular, a significant fraction of the effort during sea trials is devoted to fault diagnosis, telemetry interpretation, mission adjustment, and coordination across heterogeneous assets and operators. We argue that LLMs may be useful at this higher operational level, especially when grounded in structured vehicle state, alarms, logs, mission context, and existing planning or validation tools. At the same time, current limitations in grounding, consistency, and causal reasoning make these systems lack the predictability required to guarantee direct control, especially in real-world critical applications, although we do not exclude a future evolution in this direction. This viewpoint suggests a layered architecture in which conventional autonomy remains responsible for control, perception, estimation, and validated mission execution, whereas LLM-based modules serve as an additional support layer for debugging, operator interaction, constrained replanning support, orchestration, and post-mission analysis.
\end{abstract}

\section{Layered intelligence under field constraints}
\label{sect:introduction}
Underwater robotic systems are developed and operated under severe environmental and operational constraints, including low-bandwidth communication, delayed and missing measurements, uncertain localization, platform heterogeneity, and limited opportunities for intervention once a mission is underway. These constraints affect the design of autonomy algorithms, but they also affect the practical execution of field operations. In many sea trials, the main bottleneck is not the absence of a planner or controller, but the time needed to understand the current system state, identify the cause of degraded behavior, and decide whether the mission should continue, terminate, or be modified.

This extended abstract focuses on a specific and rather conservative question: whether LLMs can reduce part of this operational burden without being entrusted with control authority. Our main premise is that this is the more relevant question for near-term deployments. While recent work on LLM-based AI has often emphasized general planning or end-to-end autonomy, many of the costly failures in sea trials arise at a different level, namely in the interpretation of system behavior, the reconstruction of recent events, and the coordination of operators and assets under time pressure. From this perspective, LLMs and conventional autonomy modules address different problems and should be integrated and evaluated accordingly. This emphasis on bounded assistance rather than authority is also consistent with broader findings in human--robot interaction, where trust and acceptance depend strongly on how system competence, limitations, and decision processes are exposed to the user~\cite{PAH-DRB-KES-JYCC-EJdV-RP:2011,RHW-AT-JJB:2017}.

We argue that, currently, LLMs belong primarily in an operator-facing layer and, to a limited extent, at the interface with mission management. We envisage a dual use: at a command-and-control centre, where the main objective is to scrutinize data received in real time and logs after vehicle recovery; and on board the vehicles, where they may act as supervisors and help identify the most useful information to transmit, thus coping better with bandwidth limitations. In these roles, they may aggregate structured robot state, alarms, logs, planner outputs, and operator annotations into reviewable summaries, explanations, fault hypotheses, and candidate plan changes. This may also contribute to autonomy explainability through the scrutiny of data and feedback. Such explainability, hard to achieve especially in communication-limited environments, is a necessary step toward building the trust required by end users to operate the next generation of autonomous maritime unmanned systems. Any action affecting a vehicle should still pass through the existing autonomy stack together with the corresponding safety checks and mission constraints. Table~\ref{tab:layered-stack} summarizes this interpretation.

\begin{table*}[t]
\centering
\caption{Layered interpretation of underwater robotic intelligence and the corresponding role of LLM-based methods.}
\label{tab:layered-stack}
\scriptsize
\setlength{\tabcolsep}{3pt}
\begin{tabular*}{\textwidth}{@{}p{2.8cm}p{5.1cm}p{4.7cm}p{4.8cm}@{}}
\hline
\textbf{Layer} & \textbf{Typical functions} & \textbf{Representative methods} & \textbf{Reliability requirement and role of LLMs} \\
\hline
Control and safety & Stabilization, guidance, trajectory tracking, inner-loop supervision, safety interlocks & PID and model-based control, robust or adaptive control, fault protection & Very high reliability and determinism; not a suitable role for current LLMs \\
Perception & Target or obstacle detection, feature extraction, environmental interpretation, payload processing & Signal processing, learned perception, sonar pipelines, classification and detection modules & Strong sensor grounding required; LLMs may explain outputs but not replace core perception \\
State estimation and fusion & Localization, tracking, navigation filtering, multi-sensor fusion, cooperative estimation & Kalman and particle filters, SLAM, map-based localization, distributed estimation & Very high reliability and temporal consistency; LLMs should not perform core estimation \\
Communications and mission execution & Link management, task sequencing, abort logic, validated mission execution, health-triggered fallbacks & Protocol logic, behavior trees, mission managers, planners, rule-based execution monitors & High reliability; LLMs may support interpretation but should not manage communications or execute missions directly \\
Replanning and coordination & Comparison of feasible alternatives, mixed-asset tasking, planner queries, constraint handling, communication-aware coordination & Constrained planners, schedulers, optimization, rule-based coordination, human review workflows & High reliability with tool support and human review; useful for summarization, option comparison, and trade-off explanation \\
Operator support and post-mission analysis & Telemetry interpretation, fault diagnosis, status explanation, chronology reconstruction, reporting & Dashboards, log parsers, retrieval systems, analytics, tool-using LLMs & Reviewable outputs; strongest near-term use case for LLMs \\
Human--system interface & Natural-language status queries, intent translation, checklist assistance, customer-facing explanation & Structured prompting, retrieval augmentation, tool use, interfaces to autonomy services & Useful if tightly bounded by structured state and available tools; should remain advisory rather than authoritative \\
\hline
\end{tabular*}
\end{table*}

\section{Operational use cases and limitations}
Among the possible applications, fault diagnosis appears to be the most compelling. In many deployments, faults are not catastrophic but ambiguous. For instance, a payload process may restart, acoustic communication quality may degrade, or DVL bottom lock may be intermittently lost. Each event may be manageable in isolation, but their combination can reduce confidence in navigation or mission execution. The difficulty for the operator is then to reconstruct, from distributed and time-asynchronous information, which subsystem changed first, which effects were secondary, and which checks are most useful before the next deployment. An LLM that is grounded in machine-readable logs, alarms, and mission state may assist with this reconstruction by generating concise chronologies of events, identifying relevant state transitions, and ranking plausible fault hypotheses. The intended benefit is not automatic recovery, but a reduction in time-to-diagnosis.

A closely related application is telemetry and log interpretation. Underwater platforms and payloads produce large amounts of status information, often at multiple rates and across different interfaces. This information is essential, but it is not always presented at the right level of abstraction for rapid operational decisions. In particular, operators often need a short account of what changed, what remains nominal, and which subsystem is limiting the mission. This functionality may be particularly useful in multi-asset settings involving AUVs, surface gateways, and shore services. Mission replanning is also relevant, though in a more constrained form. In field conditions, replanning is rarely an unconstrained search problem; rather, it is typically a choice among a small number of feasible alternatives determined by vehicle health, communication windows, environmental conditions, and mission objectives. In this setting, an LLM-based module may be useful as an interface to the planning stack, provided that final decisions remain with the operator and the validated planning tools. More generally, this points toward LLM use patterns based on explicit tool invocation and structured intermediate representations, rather than free-form command generation~\cite{TS-JDY-RD-RR-ML-EH-LZ-NC-TS:2023}.

The previous discussion does not imply that current LLMs are broadly reliable in underwater operations. Their well-known limitations remain particularly relevant in this domain. Hallucinations, inconsistent outputs, weak grounding, and difficulty with causal and temporal reasoning become more serious when the relevant events are asynchronous, partially observed, and distributed across multiple software and hardware components, as also noted in recent surveys of language-generation failures~\cite{ZJ-NL-RF-TY-DS-YX-EI-YJB-AM-PF:2023}. For this reason, we do not presently view LLMs as suitable for direct control, low-level mission execution, or other safety-critical roles. At the same time, we do not exclude future developments in this direction. Emerging approaches for structured tool use, constrained inference, and safer decision support, e.g., systems such as TypeSafe AI's Jev~\cite{jev}, may help make this question worth revisiting. 

\section{Concluding remarks}
In summary, this extended abstract argues for a layered interpretation of robotic intelligence in underwater systems and for a correspondingly restricted role of LLMs. Our main claim is that the near-term value of these models is more likely to be found in operational support, especially fault diagnosis, telemetry interpretation, constrained replanning support, operator interaction, and post-mission analysis, than in direct autonomous control. A useful research direction, in our opinion, is therefore not just to replace the autonomy stack with LLM-based methods, but to determine how such methods can be integrated with structured system state, constrained by existing autonomy and safety mechanisms, and evaluated using field-relevant metrics such as time-to-diagnosis, operator workload, and quality of decision support.

\bibliographystyle{unsrtnat}
\bibliography{references}

@misc{jev,
  author       = {{TypeSafe AI}},
  title        = {Introduction},
  howpublished = {\emph{TypeSafe AI Documentation}. Available: \url{https://docs.typesafe.ai/introduction}},
  note         = {Accessed: 2026-10-12},
  year         = {2026}
}

@article{PAH-DRB-KES-JYCC-EJdV-RP:2011,
  author  = {Peter A. Hancock and Deborah R. Billings and Kristin E. Schaefer and Jessie Y. C. Chen and Ewart J. de Visser and Raja Parasuraman},
  title   = {A Meta-Analysis of Factors Affecting Trust in Human-Robot Interaction},
  journal = {Human Factors: The Journal of the Human Factors and Ergonomics Society},
  volume  = {53},
  number  = {5},
  pages   = {517--527},
  year    = {2011},
  doi     = {10.1177/0018720811417254}
}

@article{ZJ-NL-RF-TY-DS-YX-EI-YJB-AM-PF:2023,
  author  = {Ziwei Ji and Nayeon Lee and Rita Frieske and Tiezheng Yu and Dan Su and Yan Xu and Etsuko Ishii and Ye Jin Bang and Andrea Madotto and Pascale Fung},
  title   = {Survey of Hallucination in Natural Language Generation},
  journal = {ACM Computing Surveys},
  volume  = {55},
  number  = {12},
  pages   = {1--38},
  year    = {2023},
  doi     = {10.1145/3571730}
}

@inproceedings{RHW-AT-JJB:2017,
  author    = {Robert H. Wortham and Andreas Theodorou and Joanna J. Bryson},
  title     = {Improving Robot Transparency: Real-Time Visualisation of Robot AI Substantially Improves Understanding in Naive Observers},
  booktitle = {Proceedings of the 26th IEEE International Symposium on Robot and Human Interactive Communication (RO-MAN)},
  pages     = {1424--1431},
  year      = {2017},
  doi       = {10.1109/ROMAN.2017.8172491}
}

@inproceedings{TS-JDY-RD-RR-ML-EH-LZ-NC-TS:2023,
  author    = {Timo Schick and Jane Dwivedi-Yu and Roberto Dess\`i and Roberta Raileanu and Maria Lomeli and Eric Hambro and Luke Zettlemoyer and Nicola Cancedda and Thomas Scialom},
  title     = {Toolformer: Language Models Can Teach Themselves to Use Tools},
  booktitle = {Advances in Neural Information Processing Systems 36},
  pages     = {68539--68551},
  year      = {2023},
  doi       = {10.52202/075280-2997}
}

\end{document}